\pdfoutput=1

\documentclass[11pt]{article}

\usepackage[final]{acl}

\usepackage{times}
\usepackage{latexsym}

\usepackage[T1]{fontenc}
\usepackage{amsmath}

\usepackage[utf8]{inputenc}
\usepackage{pifont}
\usepackage{microtype}
\usepackage{graphicx}
\usepackage{subfigure}
\usepackage{booktabs} 
\usepackage{graphicx}  
\usepackage{float}     

\usepackage{multirow}
\usepackage{verbatim}
\usepackage{graphicx}
\usepackage{pifont}
\usepackage{adjustbox}
\usepackage{enumitem}
\usepackage{amsmath,amsfonts}
\usepackage{algorithmic}
\usepackage{algorithm}
\usepackage{array}
\usepackage{caption}

\usepackage{amsmath}
\usepackage{amssymb}
\usepackage{mathtools}
\usepackage{amsthm}

\usepackage{inconsolata}

\usepackage{graphicx}

\usepackage{microtype}
\usepackage{hyperref}
\usepackage{tabularx} 

\usepackage{times}
\usepackage{latexsym}
\usepackage{subcaption}
\usepackage{graphicx}
\usepackage{amsmath}
\usepackage{float}
\usepackage{booktabs} 
\usepackage[T1]{fontenc}

\usepackage[utf8]{inputenc}

\usepackage{microtype}
\usepackage{microtype}
\usepackage{hyperref}
\usepackage{inconsolata}

\usepackage{graphicx}

\title{ChartReasoner: Code-Driven Modality Bridging for Long-Chain Reasoning in Chart Question Answering}

\author{
  Caijun Jia$^{2,4}$\thanks{Equal Contribution.}, 
  Nan Xu$^{1,3}$\footnotemark[1], 
  Jingxuan Wei$^{2,4}$, 
  Qingli Wang$^{3}$, 
  Lei Wang$^{1,3}$, 
  Bihui Yu$^{2,4}$, 
  Junnan Zhu$^{1}$\thanks{Corresponding Author.}
  \\
  $^1$ MAIS, Institute of Automation, Chinese Academy of Sciences \\
  $^2$ Shenyang Institute of Computing Technology, Chinese Academy of Sciences \\
  $^3$ Beijing Wenge Technology Co., Ltd. \\
  $^4$ University of Chinese Academy of Sciences \\
  {\texttt weijingxuan20@mails.ucas.edu.cn, nan.xu@wenge.com, junnan.zhu@nlpr.ia.ac.cn}\\
}

\begin{document}
\maketitle
\begin{abstract}

Recently, large language models have shown remarkable reasoning capabilities through long-chain reasoning before responding. However, how to extend this capability to visual reasoning tasks remains an open challenge. Existing multimodal reasoning approaches transfer such visual reasoning task into textual reasoning task via several image-to-text conversions, which often lose critical structural and semantic information embedded in visualizations, especially for tasks like chart question answering that require a large amount of visual details. To bridge this gap, we propose ChartReasoner, a code-driven novel two-stage framework designed to enable precise, interpretable reasoning over charts. We first train a high-fidelity model to convert diverse chart images into structured ECharts codes, preserving both layout and data semantics as lossless as possible. Then, we design a general chart reasoning data synthesis pipeline, which leverages this pretrained transport model to automatically and scalably generate chart reasoning trajectories and utilizes a code validator to filter out low-quality samples. Finally, we train the final multimodal model using a combination of supervised fine-tuning and reinforcement learning on our synthesized chart reasoning dataset and experimental results on four public benchmarks clearly demonstrate the effectiveness of our proposed ChartReasoner. It can preserve the original details of the charts as much as possible and perform comparably with state-of-the-art open-source models while using fewer parameters, approaching the performance of proprietary systems like GPT-4o in out-of-domain settings. 

\end{abstract}

\section{Introduction}

Chart question answering (ChartQA) aims to enable models to understand and reason over structured visualizations such as bar and line charts. Recent models have improved visual-text alignment \citep{masry2023unichart, liu2022matcha}, while ChartLlama \citep{han2023chartllama} and ChartSFT \citep{meng2024chartassisstant} introduce chain-of-thought (CoT) prompting for multi-step reasoning. However, most existing ChartQA models still lack true reasoning capabilities. The CoT reasoning they produce is often shallow and short, resulting in superficial reasoning without genuine logical depth.

Large language models (LLMs) have achieved remarkable success in text-based long-chain reasoning, producing highly accurate, structured, and multi-step solutions to complex problems. This is exemplified by LLMs such as DeepSeek-R1 \citep{guo2025deepseek}. These models decompose complex problems into logical sequential steps, each building upon previous deductions to reach well-justified conclusions. However, this reasoning capability remains largely confined to the textual domain, creating a significant gap when applied to visual chart interpretation tasks. In particular, ChartQA poses unique challenges that cannot be addressed by simple extensions of text-based reasoning. Existing multimodal approaches typically convert visual inputs into textual representations via image-to-text pipelines. While effective in some contexts, this strategy often fails to preserve the structural and semantic fidelity of the original visualizations. As a result, critical layout, spatial, and quantitative details necessary for accurate reasoning in ChartQA are frequently lost, severely limiting the effectiveness of current models in this domain.

Recent advances in multimodal reasoning extend structured thinking from text to vision by converting images into textual representations to enable CoT reasoning. Methods such as R1-OneVision \citep{chen2025r1v} translate visual scenes into formal text, while R1-V \citep{chen2025r1v}, Curr-ReFT \citep{deng2025boosting}, LMM-R1 \citep{peng2025lmm} and MMEureka \citep{meng2025mm} leverage reinforcement learning to enhance object-centric and long chain reasoning. Despite these innovations, visual content is often treated as auxiliary, serialized into language at the cost of losing fine-grained cues. Local structures, color semantics, and spatial layouts are frequently abstracted or compressed. This lossy transformation undermines tasks that require precise visual grounding, such as ChartQA or scientific diagram analysis.

To address the challenges of chart-based understanding and long-chain reasoning, we propose ChartReasoner, a code-driven, two-stage framework that enhances the reasoning capabilities of multimodal large language models (MLLMs). In the first stage, we train Chart2Code, a high-accuracy model that translates diverse chart images into structured ECharts code, faithfully preserving both visual layout and underlying data semantics. This symbolic representation serves as the foundation for reasoning, bridging the visual–textual modality gap. In the second stage, we construct the ChartThink dataset by applying Chart2Code to various benchmarks, yielding 140K multi-step reasoning samples. We then train the final ChartReasoner model through supervised fine-tuning (SFT) and reinforcement learning (RL) to improve reasoning accuracy, consistency, and interpretability. This structured pipeline enables precise, scalable, and logically grounded ChartQA.

Contributions are as follows:
\begin{itemize}
    \item We introduce ChartThink, the first large-scale chart reasoning dataset with over 140K multi-step samples, supporting interpretable and logic-driven analysis across diverse chart types. We also construct Chart2Code, a dataset of 110K synthetic charts paired with accurate ECharts code, bridging the visual–textual modality gap.

    \item We propose ChartReasoner, a two-stage model that first translates chart images into symbolic ECharts code using Chart2Code, then performs multi-step reasoning on the structured representation. This improves accuracy and generalization across ChartQA tasks.
    
    \item ChartReasoner is comparable to state-of-the-art open-source models on ChartQA, ChartBench, EvoChart-QA, and ChartQAPro, while using fewer parameters, and approaches the performance of proprietary systems like GPT-4o in out-of-domain settings.
    
\end{itemize}

\section{Related Work}

\paragraph{ChartQA.}

To enhance MLLMs' chart understanding, datasets like ChartQA and others have been proposed \citep{kahou2017figureqa,kafle2018dvqa,methani2020plotqa,chaudhry2020leaf,masry2022chartqa}, covering diverse chart types and visual reasoning tasks. However, most focus on single-value or label-based answers and lack support for complex multi-step reasoning. Recent efforts such as ChartX and related work \citep{xia2024chartx,ahmed2023realcqa,masry2023unichart,meng2024chartassisstant,huang2025evochart} scale up data via synthetic generation, template-based QA, and CoT annotations, though reasoning depth remains limited. On the model side, compact models like ChartReader and others \citep{cheng2023chartreader,liu2022matcha,baechler2024screenai,masry2023unichart} show strong results on early benchmarks. LLaVA-based models including ChartLlama and its variants \citep{han2023chartllama,carbune2024chart,masry2024chartinstruct,meng2024chartassisstant,zhang2024tinychart} further enhance multimodal alignment. More recently, generalist vision-language models like Phi-3 Vision \citep{chen2024expanding} have also achieved promising performance. Nevertheless, current models still struggle with long-chain reasoning involving multi-cue integration and numerical-logical inference.

\begin{figure*}[t]
\setlength{\abovecaptionskip}{0.12cm}    
\setlength{\belowcaptionskip}{-0.3cm}
    \centering
    \includegraphics[width=\linewidth]{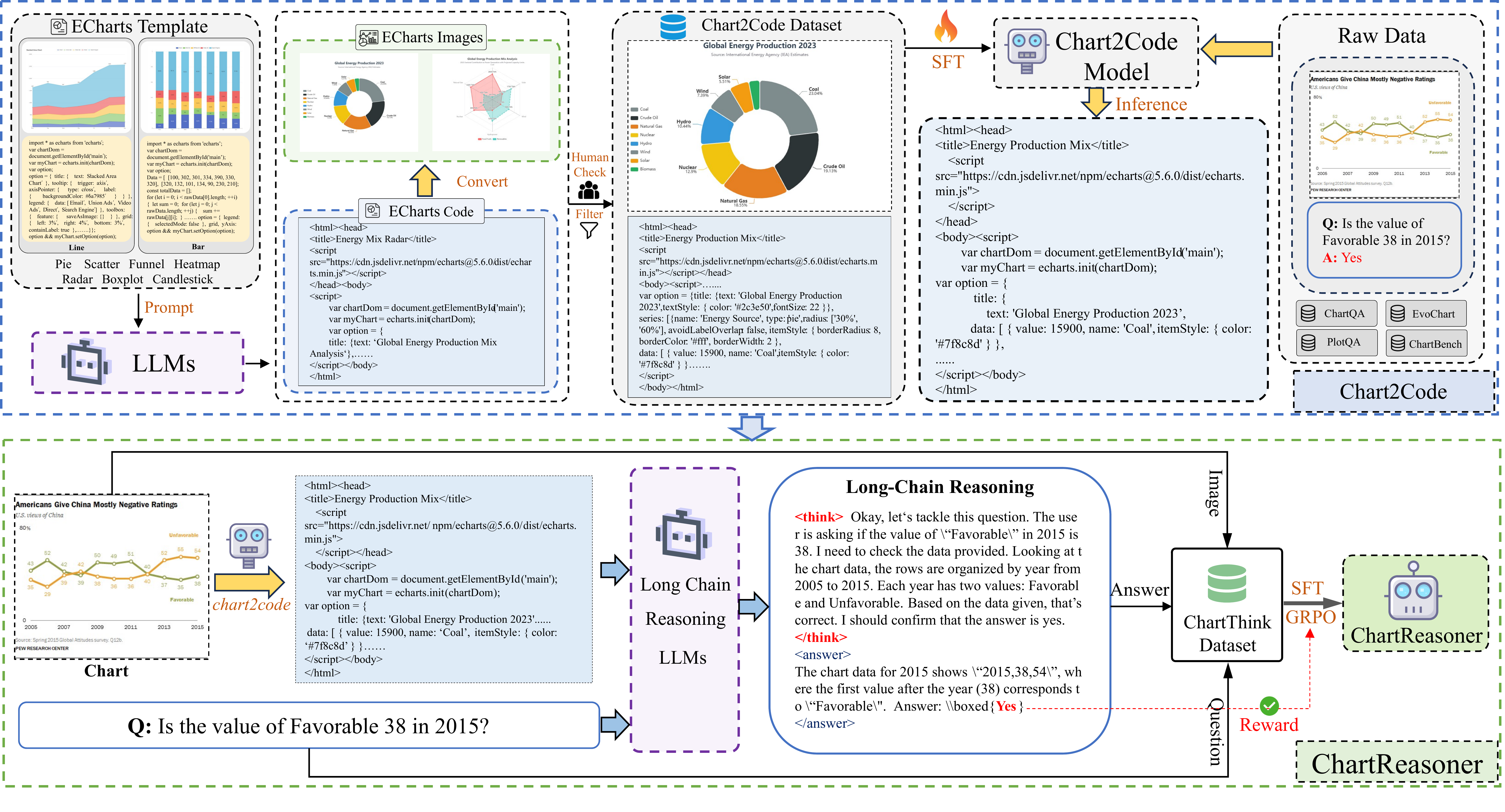}
    \caption{Overview of the data construction pipeline and model training. (1) Chart2Code: We generate a synthetic dataset by rendering chart specifications into paired images and structured code using a prompt-based pipeline. A hybrid filtering strategy is applied to ensure data quality. A vision-language model is then fine-tuned on these image–code pairs. (2) ChartReasoner: Existing ChartQA datasets are converted into structured code using the trained Chart2Code model. Reasoning traces are then generated over code representations to build a symbolic reasoning dataset. The final model is trained in two stages.}
    \vspace{-3mm}
    \label{fig:method}
\end{figure*}

\paragraph{Multimodal Long-Chain Reasoning.}

Long-chain reasoning has gained momentum in NLP with the emergence of DeepSeek-R1 \citep{guo2025deepseek}, which emphasizes structured intermediate steps. This paradigm has been extended to vision-language models (VLMs) through works like R1-OneVision \citep{yang2025r1} and Vision-R1 \citep{huang2025vision}, which convert images into formal textual representations to enable multimodal CoT training. R1-V \citep{chen2025r1v} applies Group Relative Policy Optimization (GRPO) \citep{shao2024deepseekmath} to object counting, demonstrating that small models can outperform larger ones with effective reinforcement learning. VisualThinker-R1-Zero \citep{zhou2025r1}, Curr-ReFT \citep{deng2025boosting}, LMM-R1 \citep{peng2025lmm}, and MMEureka \citep{meng2025mm} further explore RL-driven reasoning, revealing “visual aha moments” where longer outputs indicate deeper reasoning. Despite these advances, most existing methods focus on natural or generic visual inputs and struggle to handle structured representations such as charts.

To address the challenges of chart understanding and long-chain reasoning, we propose ChartReasoner, a code-driven, two-stage framework that enhances the reasoning capabilities of MLLMs. In stage one, Chart2Code translates diverse chart images into structured ECharts code, preserving visual layouts and data semantics. Compared with prior methods such as ChartMimic \citep{yang2024chartmimic}, Plot2Code \citep{wu2024plot2code}, and ChartX \citep{xia2024chartx}, which focus on layout heuristics or require costly supervision \citep{li2025metalmultiagentframeworkchart, zhang2025enhancing}, Chart2Code offers greater flexibility and fidelity via executable, symbolic code. While ChartCoder \citep{zhao2025chartcoder} adopts a similar code-centric approach, its reliance on fixed templates limits generalization. In stage two, we generate 140K multi-step reasoning samples by applying Chart2Code to existing ChartQA benchmarks, forming the ChartThink dataset. ChartReasoner is then trained with supervised and reinforcement learning, yielding accurate, consistent, and interpretable reasoning. This structured pipeline bridges the visual–textual gap and supports logically grounded ChartQA.

\section{Our Method}

\begin{figure*}[t]
\setlength{\abovecaptionskip}{0.12cm}    
\setlength{\belowcaptionskip}{-0.3cm}
    \centering
    \includegraphics[width=\linewidth]{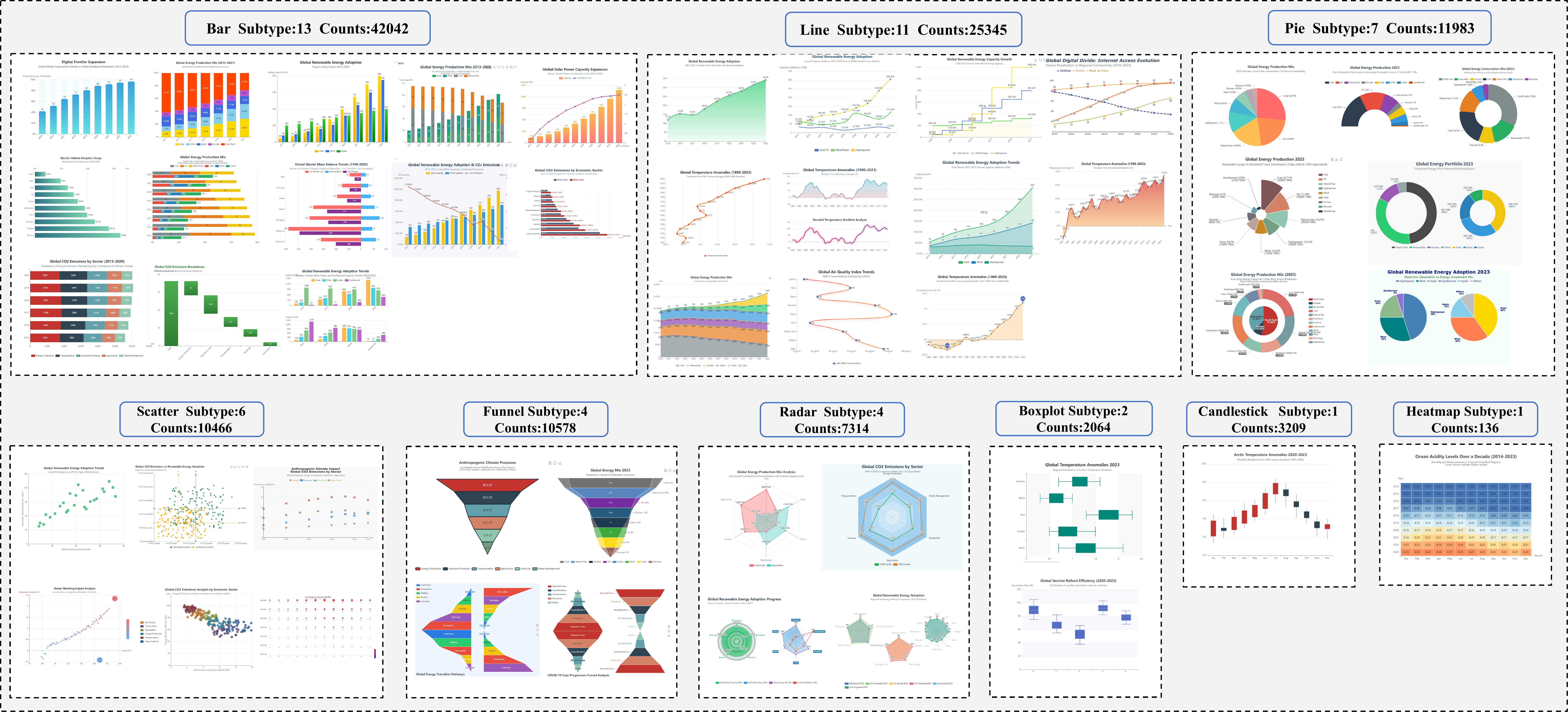}
    \caption{Statistics and distribution of chart types and subtypes in the Chart2Code dataset.}
    \vspace{-3mm}
    \label{fig:charttype}
\end{figure*}

Understanding charts remains a core challenge for MLLMs due to the gap between visual inputs and symbolic semantics. To bridge this gap, we propose a code-driven, two-stage framework that integrates visual perception with symbolic reasoning through structured chart representations. In the first stage, we introduce Chart2Code, a model that converts chart images into executable ECharts code, preserving both visual layout and semantic structure. To train it, we generate a 110K synthetic dataset via a prompt-based pipeline, rendering chart specifications into image–code pairs and applying a hybrid filtering strategy to ensure quality. The model is fine-tuned with a frozen visual encoder and a trainable decoder for accurate symbolic generation. In the second stage, we use Chart2Code to construct ChartThink, a 140K dataset of multi-step reasoning samples. These are derived by extracting structured code from existing ChartQA benchmarks and prompting a language model to generate chain-of-thought reasoning over the code. This symbolic abstraction enables precise and lossless reasoning over chart semantics. Our final model, ChartReasoner, is trained in two phases: supervised fine-tuning to establish logical competence, followed by reinforcement learning with rule-guided rewards to deepen reasoning capabilities. Overall, our framework treats code as a compositional and interpretable bridge between vision and language, enabling accurate and logically grounded chart understanding, as illustrated in \autoref{fig:method}.

\subsection{Chart2Code}
\paragraph{Echarts-format Chart Generation.}
We begin with a template library $\mathcal{T} = \{T_1, T_2, \dots, T_K\}$, which spans $K$ chart templates across 9 major categories and 49 subtypes. For each template $T_k \in \mathcal{T}$, we construct a prompt $p_k$ to guide an LLM in generating diverse ECharts code. The detailed prompt design is provided in \autoref{appendix:html_pro}. The ECharts code $c_j$ for a sample $j$ is obtained as:

{\small
\begin{equation}
c_{j} = G_{\text{LLM}}(p_k),
\end{equation}
}

where $G_{\text{LLM}}(\cdot)$ denotes the LLM that produces structured chart code given the input prompt.

\paragraph{Quality Filtering Pipeline.}

The generated ECharts code is rendered into images and subjected to a rigorous quality control process. We combine automated pixel-level filtering with manual review to enhance image quality. In the automated stage, each image is converted to the Hue-Saturation-Value color space to extract saturation and brightness features and is downsampled to reduce computational overhead. Blank and noisy images are then removed using sparse content detection and white-background noise filtering. In the manual stage, we further eliminate edge cases that are difficult to detect automatically. As a result, we retain approximately 110K high-quality charts from the initial set. Detailed distribution statistics are provided in \autoref{fig:charttype}, covering 9 major categories and 49 subcategories. Specific data examples are included in the \autoref{appendix:c2c_case}.

\paragraph{Chart2Code Model.}

To enable high-fidelity chart reconstruction from images, we construct a large-scale chart-to-code dataset and use it to train a multimodal model capable of translating chart images into their corresponding ECharts code. We fine-tune a vision-language model on this dataset for the chart-to-code generation task. Given a chart image $x_i$, the model predicts its corresponding ECharts code sequence $\mathbf{c_i} = (c_{i,1}, c_{i,2}, \dots, c_{i,L_i})$, where $L_i$ is the token length of the code. The model is trained to maximize the likelihood of the target sequence conditioned on the input image. The model parameters are denoted as $\theta = {\theta_{\text{VE}}, \theta_{\text{LD}}}$, where $\theta_{\text{VE}}$ refers to the visual encoder (frozen during training), and $\theta_{\text{LD}}$ denotes the language decoder parameters. The training objective is to minimize the loss function $\mathcal{L}_{\text{C2C}}$:

{\small%
\begin{equation}
\mathcal{L}_{\text{C2C}}(\theta_{\text{LD}}) = - \sum_{i=1}^{N_{\text{C2C}}} \sum_{t=1}^{L_i} \log P(c_{i,t} \mid x_i, \mathbf{c}_{i,<t}; \theta)
\end{equation}
}%
where $\mathbf{c}_{i,<t}$ represents the sequence of previously generated (ground-truth) tokens $(c_{i,1}, \dots, c_{i,t-1})$ for the $i$-th sample, $N_{\text{C2C}}$ denotes the total number of samples in the chart-to-code dataset.

This training strategy enables the model to effectively extract both structural and semantic information from visual inputs and generate accurate, executable code.

\subsection{ChartThink Construction and Collection}
Current Chart QA datasets primarily consist of image-question-answer triplets, lacking explicit annotations of intermediate reasoning steps. This limits their effectiveness in training models that require step-by-step reasoning grounded in chart content. To address this limitation, we construct a code-driven reasoning dataset that extends traditional QA data with model-generated reasoning paths anchored in chart code. The construction pipeline is as follows.

\paragraph{ChartThink Construction.}

We begin by consolidating existing datasets into a unified collection, denoted as $\mathcal{D}_{\text{orig}} = \{ (x_k, q_k, a_k) \}_{k}$, where $x_k$ represents a chart image, $q_k$ is a corresponding question, and $a_k$ is the ground-truth answer. Each question is annotated with a reasoning type, and each chart is labeled with a structural type. To ensure broad coverage and balanced representation, we perform stratified sampling across both dimensions to select a representative subset. For each sampled chart image $x_k$, we apply the trained Chart2Code model to generate its corresponding ECharts specification $c_k$. The generated code, together with the question $q_k$, is then provided as input to a long-chain reasoning LLM, which outputs a reasoning path $r_k$ and a predicted answer $\tilde{a}_k$:

{\small
\begin{equation}
(r_{k}, \tilde{a}_k) = G_{\text{LC-R}}(\text{Prompt}(\text{Chart2Code}(x_k), q_k))
\end{equation}
}

To ensure data quality, we retain only those samples where the predicted answer $\tilde{a}_k$ exactly matches the ground-truth answer $a_k$. The final constructed dataset, referred to as ChartThink, is defined as:

{\small
\begin{equation}
\mathcal{D} = \{ (x_j, q_j, r_{j}, a_j) \}_{j=1}^{N}
\end{equation}
}

Here, $x_j, q_j, r_{j}, a_j$ represent the chart image, the corresponding question, the generated reasoning path, and the verified answer for the $j$-th sample, respectively. During training, the input to the reasoning model consists of the chart-question pair $(x_j, q_j)$, while the target output is the concatenated sequence of the reasoning path $r_j$ followed by the final answer $a_j$.

\paragraph{ChartThink Collection.}

\begin{figure}[t]
\setlength{\abovecaptionskip}{0.12cm}    
\setlength{\belowcaptionskip}{-0.3cm}
    \centering
    \includegraphics[width=\linewidth]{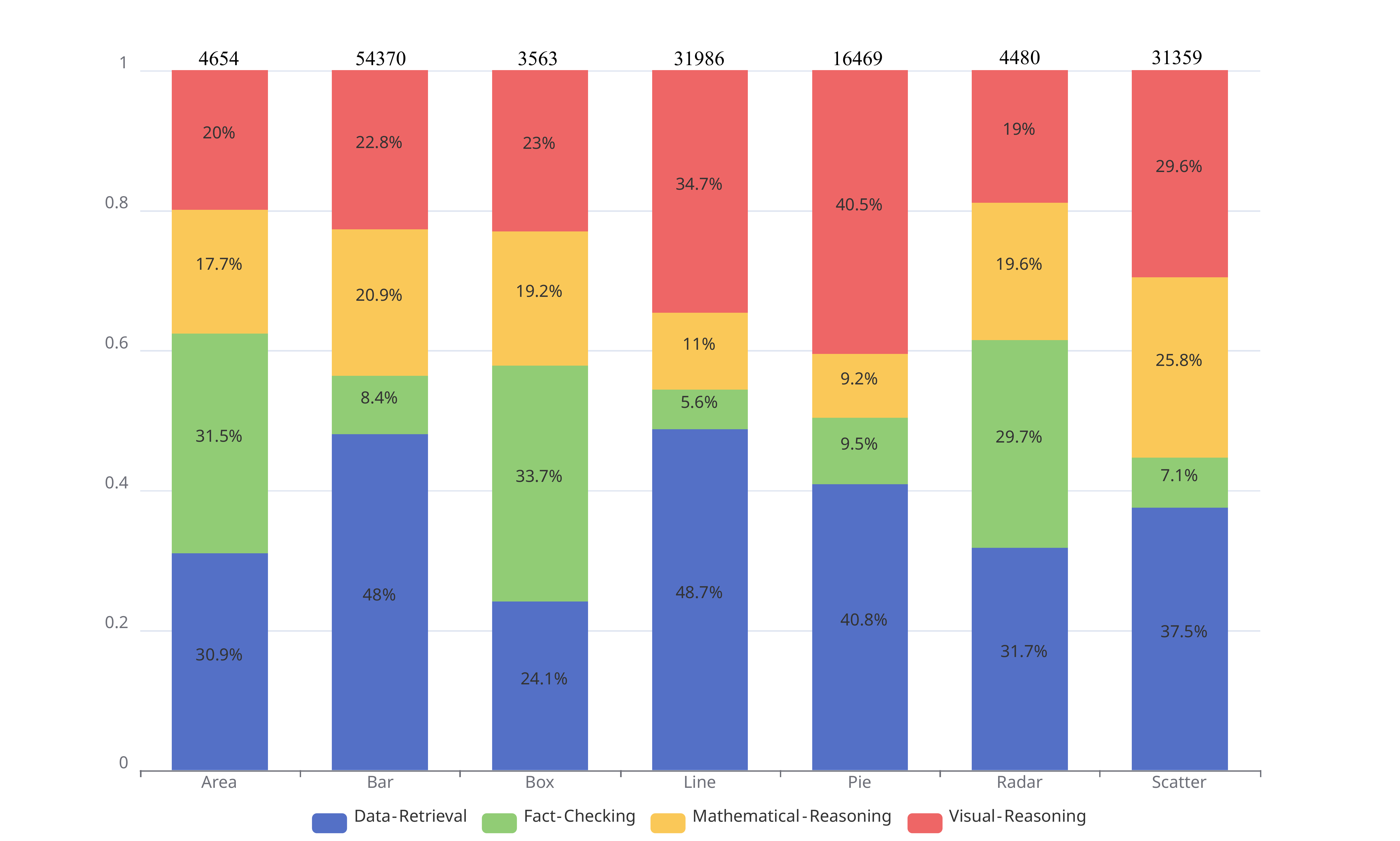}
    \caption{ChartThink dataset statistics and distribution.}
    \vspace{-3mm}
    \label{fig:cot_data}
\end{figure}

We construct the ChartThink dataset by aggregating and cleaning a wide range of existing ChartQA datasets, including ChartQA \citep{masry2022chartqa}, EvoChart \citep{huang2025evochart}, ChartBench \citep{xu2023chartbench}, and PlotQA \citep{methani2020plotqa}. These datasets collectively encompass diverse chart types and question styles commonly found in practical applications. Following the unified code-driven data pipeline introduced earlier, we systematically process all collected data to ensure consistency and correctness. After filtering out low-quality or mismatched samples, we obtain a high-quality subset containing over 140K examples, each paired with verified answers and intermediate reasoning traces. To better understand the dataset composition, we conduct a detailed analysis of the reasoning types and chart structures. \autoref{fig:cot_data} presents the ChartThink dataset statistics and its distribution over four reasoning categories and seven chart types, providing a strong foundation for training models on complex ChartQA.

\subsection{ChartReasoner}

\paragraph{Supervised Fine-Tuning.} 
The reasoning model is first trained using SFT on the $\mathcal{D}$ dataset. Given a chart image $x_j$ and a question $q_j$, the model is trained to generate a target output sequence $\mathbf{y}_j = (y_{j,1}, y_{j,2}, \dots, y_{j, K_j})$, which consists of a reasoning path followed by the final answer, and contains $K_j$ tokens. The model parameters are denoted as $\theta = \{\theta_{\text{VE}}, \theta_{\text{LD}}\}$, where $\theta_{\text{VE}}$ refers to the visual encoder (kept frozen during training), and $\theta_{\text{LD}}$ denotes the parameters of the language decoder. The SFT objective is to minimize the loss function $\mathcal{L}_{\text{SFT}}$, is defined as:

{\small%
\begin{equation}
\mathcal{L}_{\text{SFT}}(\theta_{\text{LD}}) = - \sum_{j=1}^{N} \sum_{t=1}^{K_j} \log P(y_{j,t} \mid x_j, q_j, \mathbf{y}_{j,<t}; \theta)
\end{equation}
}%
where $\mathbf{y}_{j,<t}$ represents the sequence of previously generated (ground-truth) tokens $(y_{j,1}, \dots, y_{j,t-1})$ for the $j$-th sample. 

This approach improves response uniformity and provides a stable foundation for the subsequent reinforcement learning stage.

\paragraph{Reinforcement Learning with GRPO.}
While supervised fine-tuning equips the model with fundamental chart understanding, it also reveals a common failure mode: over-generation of verbose reasoning chains, even when the input lacks sufficient information. This over-reasoning behavior compromises answer reliability. To mitigate this, we adopt a reinforcement learning phase using GRPO. Unlike standard policy optimization methods such as PPO~\citep{schulman2017proximal}, GRPO generates multiple candidate responses per input and optimizes them jointly via intra-group normalization. This stabilizes training and encourages the model to favor concise and accurate outputs.

We design structured, rule-based reward functions that explicitly measure answer quality across dimensions such as factual accuracy, formatting correctness, and response length. These reward signals guide the model to suppress hallucinations and over-reasoning, promoting disciplined and generalizable reasoning. Overall, this RL phase aligns the model's outputs with practical expectations and user preferences, significantly enhancing robustness across diverse ChartQA scenarios.

\section{Experiments}

\subsection{Experimental Setup}
\paragraph{Datasets.}

To evaluate the performance of our proposed ChartReasoner on ChartQA, we conducted experiments on four representative benchmarks: ChartQA \citep{masry2022chartqa}, EvoChart-QA \citep{huang2025evochart}, ChartQAPro \citep{masry2025chartqapro}, and ChartBench \citep{xu2023chartbench}. These datasets cover a broad range of chart types and reasoning tasks, from simple visualizations to complex real-world settings involving dashboards, infographics, and multi-chart compositions. ChartQA and EvoChart-QA emphasize real-world chart understanding with fine-grained reasoning and retrieval tasks, while ChartQAPro focuses on challenging scenarios such as multi-turn, hypothetical, and unanswerable questions. ChartBench provides large-scale evaluation across diverse chart types. We also assessed the Chart2Code on EvoChart-QA to evaluate its capability to reconstruct complex charts. Further implementation details are in \autoref{appendix:imp_detail}.

\paragraph{Evaluation Metrics \& Baselines.}
For ChartQA, we follow the official protocol for each benchmark. For Chart-to-Code, we adopt execution success rate and GPT-4V \footnote{The version is gpt-4-vision-preview} visual similarity scoring (1–10), following Plot2Code \citep{wu2024plot2code}. The specific prompt is provided in \autoref{appendix:gpt_pro}.

We benchmark our model against a wide range of MLLMs, including proprietary models like Claude-3.5-Sonnet\footnote{\url{https://www.anthropic.com/news/claude-3-5-sonnet}}, Gemini-Flash-1.5/2.0 \citep{team2024gemini}, GPT-4-turbo, and GPT-4o \citep{achiam2023gpt}, as well as open-source models such as InternVL2 \citep{chen2024far}, Phi-3-Vision \citep{abdin2024phi}, LLaVA-V1.5 \citep{liu2024improved}, InternLM-XComposer \citep{dong2024internlm}, Qwen-VL \citep{bai2025qwen2, wang2024qwen2}, and CogVLM2 \citep{hong2024cogvlm2}. We also include domain-specific baselines: ChartLlama \citep{han2023chartllama}, ChartAst \citep{meng2024chartassisstant}, ChartIns \citep{masry2024chartinstruct}, ChartGemma \citep{masry2024chartgemma}, TinyChart \citep{zhang2024tinychart}, and EvoChart \citep{huang2025evochart}. For Chart-to-Code, we use ChartCoder \citep{zhao2025chartcoder} as the main baseline. To validate the structural richness of our dataset, we conduct controlled training with Qwen2.5-VL-7B \citep{bai2025qwen2} on both EvoChart and ours.

\subsection{Main Results} 

\setlength{\tabcolsep}{10pt}  
\renewcommand{\arraystretch}{1.0} 
\begin{table*}[!t]

\setlength{\abovecaptionskip}{0.12cm}    
\setlength{\belowcaptionskip}{-0.6cm}
\centering

\resizebox{0.8\textwidth}{!}{ 

\small

\begin{tabular}{lccccc}
\toprule
\textbf{Model Name} & \textbf{Size(B)} & \textbf{Evochart-QA} & \textbf{ChartQA} & \textbf{ChartBench} & \textbf{ChartQAPro} \\
\midrule
\multicolumn{6}{c}{\textbf{Closed-source}} \\
\midrule 
Claude-3.5-Sonnet    & --  & --   & \textbf{90.80}  & --    & 43.58 \\
Gemini-Flash-2.0~\citep{team2024gemini}    & --  & --   & --    & --    & \textbf{46.85} \\
Gemini-1.5-Flash~\citep{team2024gemini}    & --  & 27.90 & 79.00    & --    & 42.96 \\
Gemini-1.5-Pro~\citep{team2024gemini}      & --  & 32.20 & 87.20  & --    & -- \\
GPT-4-turbo~\citep{achiam2023gpt}         & --  & 40.30 & 62.30  & --    & -- \\
GPT-4o~\citep{achiam2023gpt}              & --  & \textbf{49.80} & 85.70  & \textbf{59.45} & 37.67 \\

\midrule
\multicolumn{6}{c}{\textbf{Open-source}} \\
\midrule 
InternVL2-Llama3~\citep{chen2024far}    & 76 & --   & \textbf{88.40}  & --    & -- \\
Qwen2-VL~\citep{wang2024qwen2}            & 72 & --   & 88.30  & --    & -- \\
Intern-VL2~\citep{chen2024far}     & 40 & \textbf{49.00}   & 86.20  & --    & -- \\
CogVLM2~\citep{hong2024cogvlm2}             & 19 & 21.90 & 81.00    & --    & -- \\
Intern-VL2~\citep{chen2024far}      & 8  & 38.60 & 81.50  & --    & -- \\
Intern-VL2.5~\citep{chen2024expanding}    & 8  & --   & 84.80    & --    & 35.67 \\
LLaVA-v1.5~\citep{liu2024improved}          & 7  & --   & 55.32 & 23.39 & -- \\
Internlm-XComp.-v2~\citep{dong2024internlm}  & 7  & --   & 72.64 & 47.78 & -- \\
QwenVL-Chat~\citep{Qwen-VL}         & 7  & 19.70 & 83.00    & 26.98 & 35.59 \\
Qwen2.5-VL~\citep{bai2025qwen2}          & 7  & 46.80 & 85.00    & \textbf{54.06} & \textbf{36.61} \\
Phi3-Vision~\citep{abdin2024phi}         & 4  & 39.50 & 81.40  & --    & 24.73 \\

\midrule
\multicolumn{6}{c}{\textbf{Chart MLLMs}} \\
\midrule 
ChartLlama~\citep{han2023chartllama}          & 13 & 9.50  & 69.66 & 21.71 & -- \\
ChartAst-S~\citep{meng2024chartassisstant}          & 13 & 12.90 & 79.90  & --    & -- \\
ChartIns-Llama2~\citep{masry2024chartinstruct}     & 7  & 16.80 & 66.64 & --    & 4.88 \\
EvoChart~\citep{huang2025evochart}            & 4  & \textbf{54.20} & 81.50  & --    & -- \\
ChartIns-FlanT5~\citep{masry2024chartinstruct}     & 3  & 24.30 & 64.20  & --    & -- \\
ChartGemma~\citep{masry2024chartgemma}          & 3  & 30.60 & 80.16 & --    & 6.84 \\
TinyChart~\citep{abdin2024phi}           & 3  & 25.50 & 83.60  & --    & 13.25 \\
\midrule 
ChartReasoner-SFT(Ours)       & 7  & 47.04& 86.76 & 55.10  & 37.94 \\
ChartReasoner-GRPO(Ours)       & 7  & 48.10& \textbf{86.93} & \textbf{55.20}  & \textbf{39.97} \\
\bottomrule

\end{tabular}
}
\caption{Comparisons of ChartReasoner and Baselines on Four ChartQA Benchmarks.}
\label{tab:chartqa_res}
\end{table*}

\paragraph{ChartQA Results.}

We comprehensively evaluate our ChartReasoner model and a wide range of baseline models, including both general-purpose MLLMs and chart-specialized models, across four benchmark datasets. Among them, ChartQA and ChartBench are in-domain datasets, while ChartQAPro and EvoChart-QA serve as out-of-domain evaluations to test generalization performance. The results are shown in \autoref{tab:chartqa_res}.

In the ChartQA benchmark, the proprietary Claude-3.5-Sonnet model achieves top-tier performance. However, our ChartReasoner significantly outperforms all open-source 7B models and surpasses the majority of chart-specialized baselines, demonstrating its strong reasoning capability in structured visual tasks. A similar trend is observed in ChartBench, where GPT-4o leads among proprietary models, yet our model achieves state-of-the-art results among open-source and domain-specific competitors. These findings confirm that while proprietary models still retain an edge on in-domain datasets, strengthening reasoning and analysis ability can bridge this gap and yield competitive results. In the EvoChart-QA benchmark, which contains long and complex real-world charts, GPT-4o shows relatively weaker performance. In contrast, EvoChart, a chart-specialized model trained on similar data, performs better but shows clear limitations on ChartQA, indicating limited cross-domain generalization due to data-specific overfitting and a smaller model scale. Notably, our ChartReasoner matches GPT-4o’s performance and outperforms its own base model Qwen2.5-VL, confirming its enhanced capacity for long-chain visual reasoning and data adaptation. Lastly, in the ChartQAPro benchmark, Gemini-Flash-2.0 stands out among proprietary models. Still, ChartReasoner surpasses even GPT-4o in this domain-shifted setting. This reveals that many proprietary models struggle with domain transfer in chart understanding, whereas ChartReasoner's consistent performance under both in-domain and out-of-domain conditions underscores the importance of improving reasoning and abstraction abilities to enhance chart-centric generalization.

Models refined with GRPO after SFT consistently outperform those trained with SFT alone across all benchmarks. GRPO enhances reasoning quality, reduces over-explanation, and promotes more structured, precise outputs, demonstrating its effectiveness in visual reasoning.

\subsection{Ablation Study} 

\paragraph{Chart-to-Code Performance Evaluation.} 

To comprehensively evaluate the effectiveness of our Chart2Code, we present the results in \autoref{tab:Chart-to-Code_Models}, which consolidates comparisons across different datasets and training scales. Specifically, we assess model performance on a real-world test set derived from EvoChart-QA, using GPT-4V visual similarity scores and pass rates as evaluation metrics.

Our Chart2Code, trained on the proposed ECharts-based dataset, significantly outperforms models trained on the EvoChart dataset under comparable training sizes. The results highlight notable improvements in both visual fidelity and pass rate, demonstrating the higher quality and diversity of our data. This suggests that our dataset enables better generalization and more accurate chart reconstruction, even for complex and diverse chart types encountered in practice.

In addition, our model trained on ECharts-based data exhibits superior performance compared to those trained on large-scale, Python-generated chart datasets. Despite the latter having access to more training examples, their performance lags behind in both robustness and reconstruction accuracy. This underscores the importance of data realism and expressiveness, qualities more inherently present in ECharts specifications, for effectively training chart generation models.

We also analyze the impact of data volume by training models on subsets of 30k, 50k, 70k, and 110k chart–code pairs. Results show that while increasing the dataset size generally improves performance, the gains begin to plateau beyond 70k examples. This saturation effect suggests that 110k samples are sufficient to maximize the model’s reconstruction capability.

\begin{table}[h]
\setlength{\abovecaptionskip}{0.12cm}    
\setlength{\belowcaptionskip}{-0.6cm}
\centering
\small 
\resizebox{\columnwidth}{!}{ 

\begin{tabular}{lcccccccc}
\toprule
\textbf{Model} & \textbf{Data} & \textbf{Similarity} & \textbf{bar} & \textbf{line} & \textbf{pie} & \textbf{scatter} &\textbf{Rate} & \textbf{Types} \\
\midrule

ChartCoder          & 160k & 3.64 & 4.18 & 3.91 & 3.25 & 3.22 & 82.40\% & 27 \\
Chart2Code-Evo. & 70k  & 3.84 & 4.63 & 4.16 & 3.94 & 2.63 & 89.10\% & 4  \\
Chart2Code & 30k  & 2.39 & 3.12 & 2.24 & 2.81 & 1.39 & 88.20\% & 49  \\
Chart2Code & 50k  & 3.62 & 4.37 & 3.81 & 4.17 & 2.13  & 90.60\% & 49 \\
Chart2Code          & 70k & 4.21 & 5.17 & 4.20 & 4.23 & 3.24 & 91.00\% & 49 \\
Chart2Code     & 110k & \textbf{4.34} & \textbf{5.26} & \textbf{4.21} & \textbf{5.12} & \textbf{3.77} & \textbf{92.40\%} & 49 \\
\bottomrule
\end{tabular}
}
\caption{Chart2Code Performance on EvoChart-QA: GPT-4V Similarity Scores (including Breakdown by Chart Types) and Overall Pass Rates.}
\label{tab:Chart-to-Code_Models}
\end{table}

\paragraph{Sensitivity Analysis}

We conduct a sensitivity analysis to evaluate how chart type affects both chart reconstruction and downstream reasoning performance. As shown in \autoref{tab:Chart-to-Code_Models}, the Chart2Code module exhibits strong performance on bar and pie charts, while its accuracy declines for scatter and line charts. Scatter plots often contain dense, overlapping points that hinder precise encoding, whereas many line charts in EvoChart are multi-series or include complex visual encodings, making them particularly challenging to parse. These characteristics, along with their relative scarcity in the training data, contribute to consistently lower reconstruction accuracy, especially for line charts, across all models.

\begin{figure}[t]
\setlength{\abovecaptionskip}{0.12cm}    
\setlength{\belowcaptionskip}{-0.3cm}
    \centering
    \includegraphics[width=\linewidth]{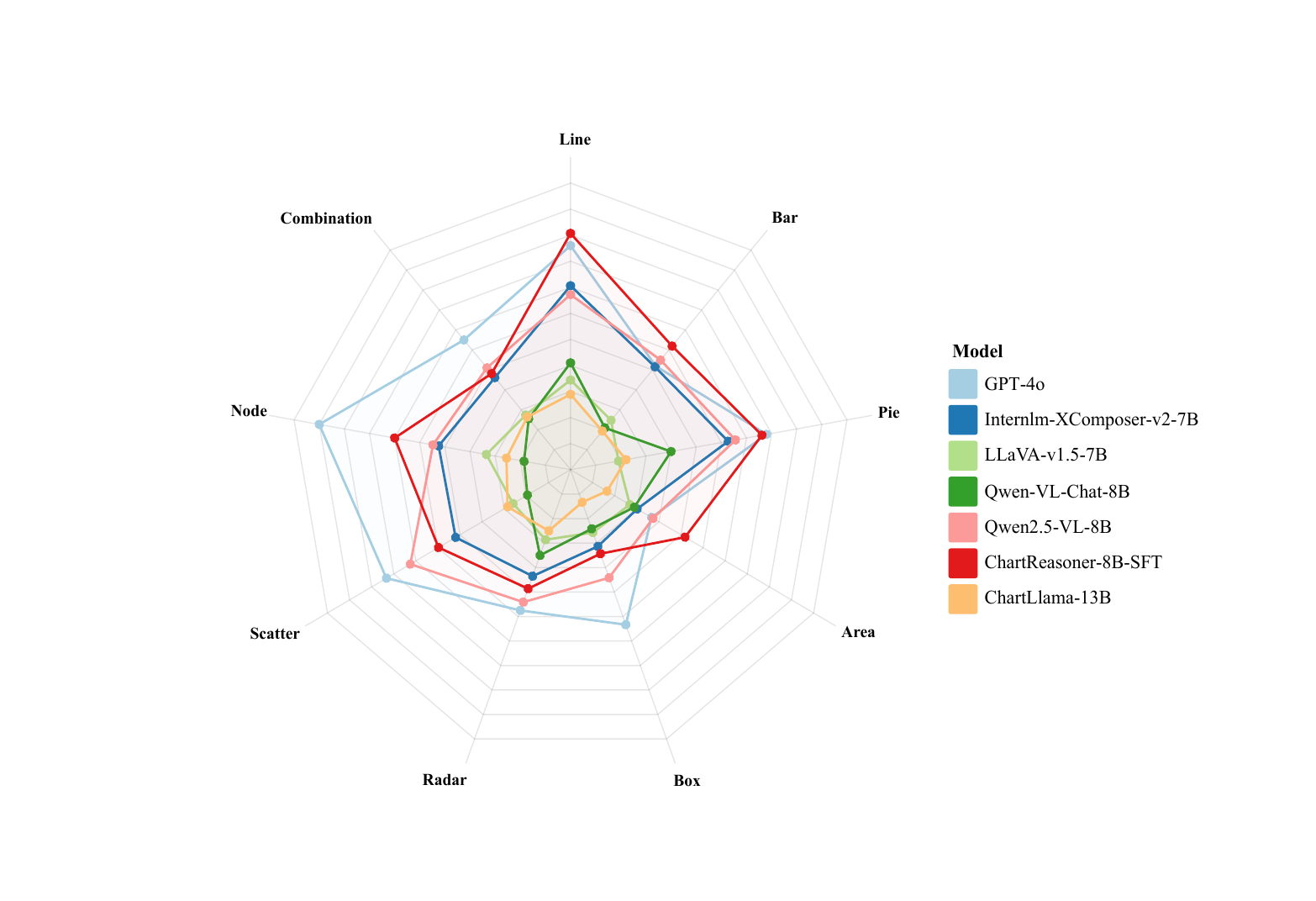}
    \caption{ChartBench Performance Across Chart Types.}
    \vspace{-3mm}
    \label{fig:ChartBench-charttype}
\end{figure}

This reconstruction quality directly influences reasoning performance in the ChartReasoner module. As shown in \autoref{fig:ChartBench-charttype} and \autoref{fig:EvoChart-charttype}, ChartReasoner achieves competitive results on bar and pie charts but underperforms on scatter, line, and box plots. Notably, the stronger results on bar, line, and pie charts within ChartBench align with their higher frequency in our training data, which enhances reconstruction robustness and, in turn, improves reasoning accuracy. These observations highlight a strong correlation between reconstruction reliability and downstream performance, underscoring the importance of both visual complexity and data distribution in ChartQA tasks.

\paragraph{Impact of Different Dataset Sources.}

We further investigate how the choice of training datasets influences downstream chart reasoning performance when generating chain-of-thought examples. To this end, we sample 20k instances from ChartQA, EvoChart, ChartBench, and PlotQA, and convert them into reasoning examples using our chart-to-code distillation pipeline. As shown in  \autoref{tab:chartreasoner-performance}, using training data that share the same distribution as the evaluation benchmark consistently leads to the best performance, highlighting the importance of distribution alignment. Among the evaluated datasets, PlotQA yields the lowest performance across all benchmarks. This outcome is likely due to its synthetic construction, limited visual variety, and narrow set of chart types, which are primarily restricted to bar, line, and dot charts. These characteristics make it less representative of real-world chart scenarios. In comparison, training data derived from EvoChart achieve better generalization, especially on EvoChart-QA and ChartQAPro. EvoChart includes a broader range of chart types, such as pie and scatter, and its charts are more visually aligned with those found in practical applications, which improves cross-domain performance. ChartBench offers strong results when evaluated on its own distribution but shows reduced effectiveness on other benchmarks, suggesting limited transferability. Overall, these findings emphasize the importance of dataset diversity when training chart reasoning models capable of robust generalization.

\begin{table}[t]
\setlength{\abovecaptionskip}{0.12cm}    
\setlength{\belowcaptionskip}{-0.6cm}
\centering
\small 
\resizebox{\columnwidth}{!}{ 
\begin{tabular}{lcccccc}
\toprule
\textbf{DataSet}   & \textbf{Evochart-QA} & \textbf{ChartQA} & \textbf{ChartBench} & \textbf{ChartQAPro} \\
\midrule
ChartQA   & 41.3 & \textbf{86.56} & 51.43 & 35.64 \\
EvoChart   & \textbf{42.8} & 85.48 & 52.38 & \textbf{36.05} \\
ChartBench  & 40.5 & 81.56 & \textbf{54.76} & 32.36 \\
PlotQA     & 40.2 & 83.00 & 47.80 & 34.69 \\
\bottomrule
\end{tabular}
}
\caption{Impact of Different Dataset Sources on Downstream Chart Reasoning Performance.}
\label{tab:chartreasoner-performance}
\end{table}

\section{Conclusion}

We present ChartReasoner, a code-driven, two-stage framework that bridges visual chart understanding and long-chain reasoning in multimodal large language models. Our approach introduces Chart2Code, which translates chart images into high-fidelity ECharts code and constructs the ChartThink dataset with over 140K multi-step reasoning samples. This enables precise, interpretable, and scalable ChartQA. Unlike previous models that rely on shallow chain-of-thought reasoning or lossy image-to-text conversions, ChartReasoner leverages structured symbolic representations to retain critical layout, semantic, and quantitative details. Extensive evaluations on four benchmarks show that ChartReasoner generalizes well and achieves strong reasoning performance. The framework supports faithful information extraction and logical inference, enabling more accurate and transparent decision-making. Our results underscore the value of symbolic representation and structured reasoning in chart-based visual understanding by tightly integrating visual parsing with logical inference. 

\clearpage

\section*{Limitations}
Our study is comprehensive but has certain limitations that we aim to address in future research. First, due to computational constraints, we conduct all experiments using a 7B-parameter model. Although this setting yields promising results, scaling to larger models may further enhance performance and generalization capabilities. Second, the current evaluation focuses primarily on benchmark-style synthetic and semi-structured charts. The generalization of our method to more complex, real-world visualizations remains an open challenge. 

\bibliography{custom}

\appendix

\section{Implementation Details.} 
\label{appendix:imp_detail}
We use DeepSeek-R1 as a controllable code and reasoning generator in both stages of data construction. We use Qwen2.5-VL-7B \citep{bai2025qwen2} as the backbone and perform supervised fine-tuning on 8 A100 80GB GPUs. The vision tower and projection layers are frozen, while the language model is fully trainable. Training runs for 4 epochs with an effective batch size of 8, using BF16 precision. We apply the AdamW \citep{loshchilov2017decoupled} optimizer with learning rate 1e-5. The maximum sequence length is 4096 tokens, and images are resized to 512×512 pixels. We further apply GRPO for 2 epochs starting from the SFT checkpoint. The model generates 8 completions per input, with reward-weighted selection based on accuracy, format correctness, and length suitability. 

\section{Qualitative Analysis}
\label{appendix:Qualitative_Analysis}
\begin{figure*}[!htbp]
    \centering
    \includegraphics[width=\linewidth]{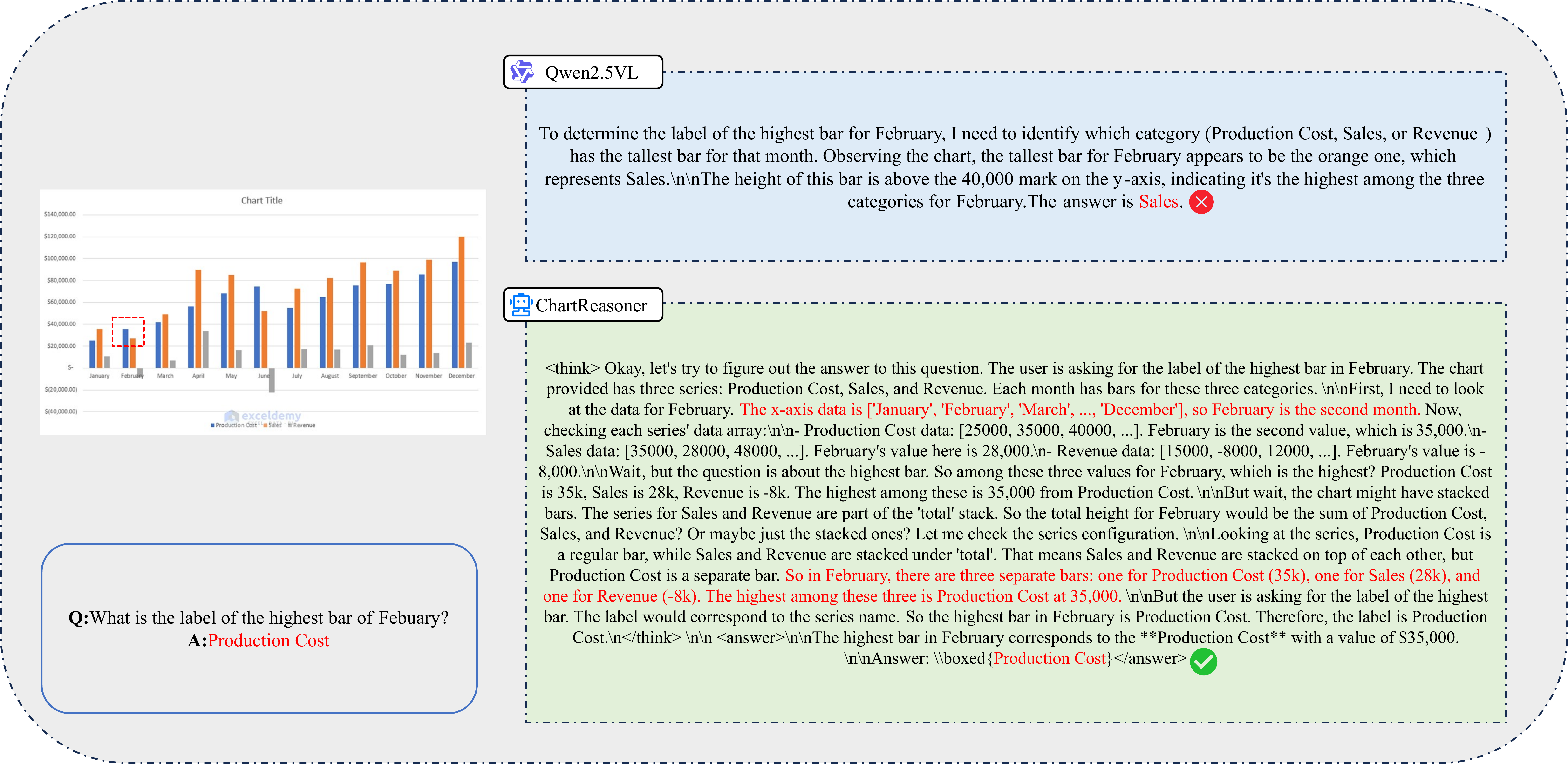}
    \caption{Comparison of Model Responses in ChartQA on a Bar Chart.}
    \vspace{-3mm}
    \label{fig:Example11}
\end{figure*}

\begin{figure*}[!htbp]
    \centering
    \includegraphics[width=\linewidth]{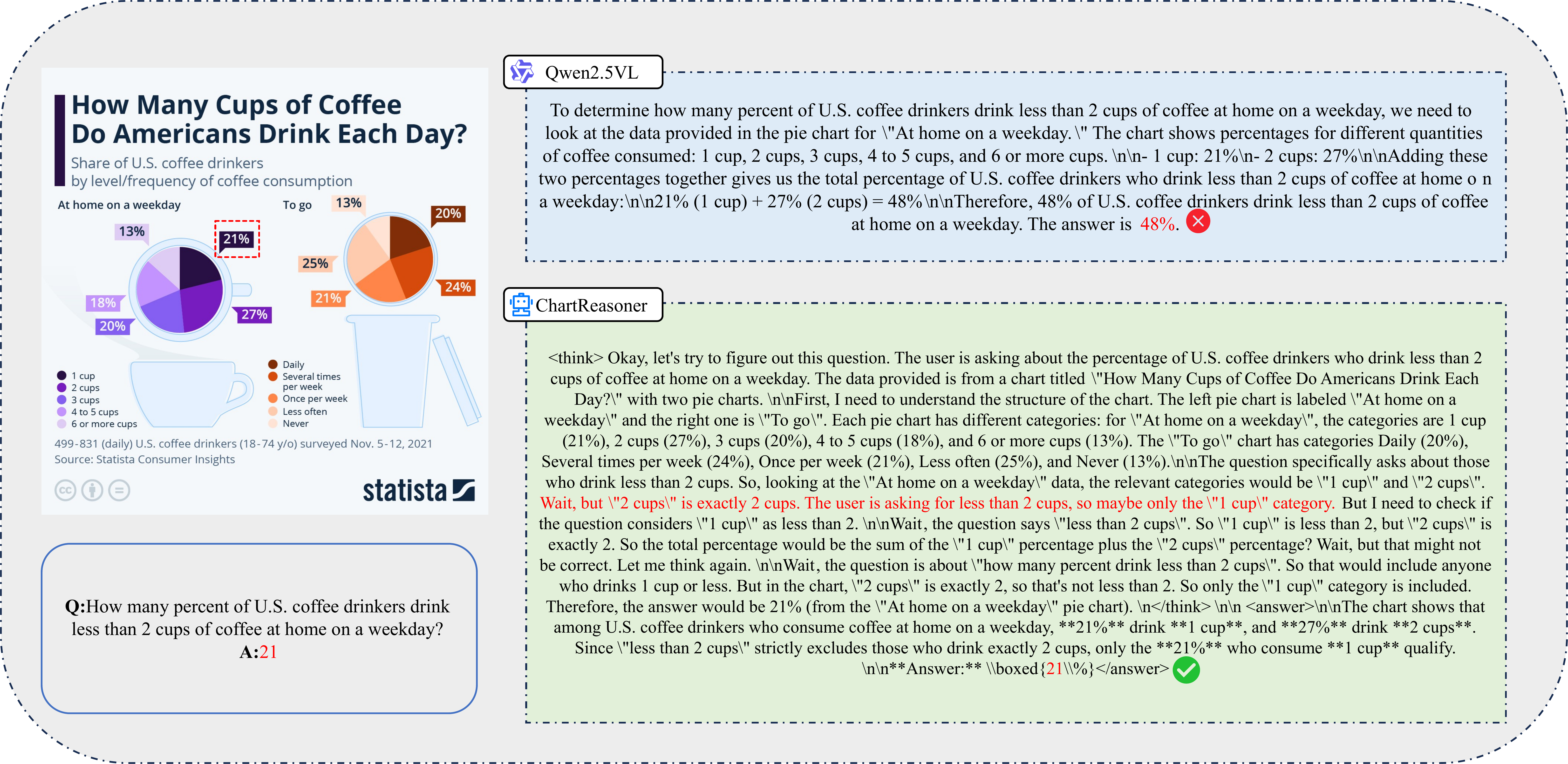}
    \caption{Comparison of Model Responses in ChartQA on a Pie Chart.}
    \label{fig:Example2}
\end{figure*}

\begin{figure*}[t]
    \centering
    \includegraphics[width=\linewidth]{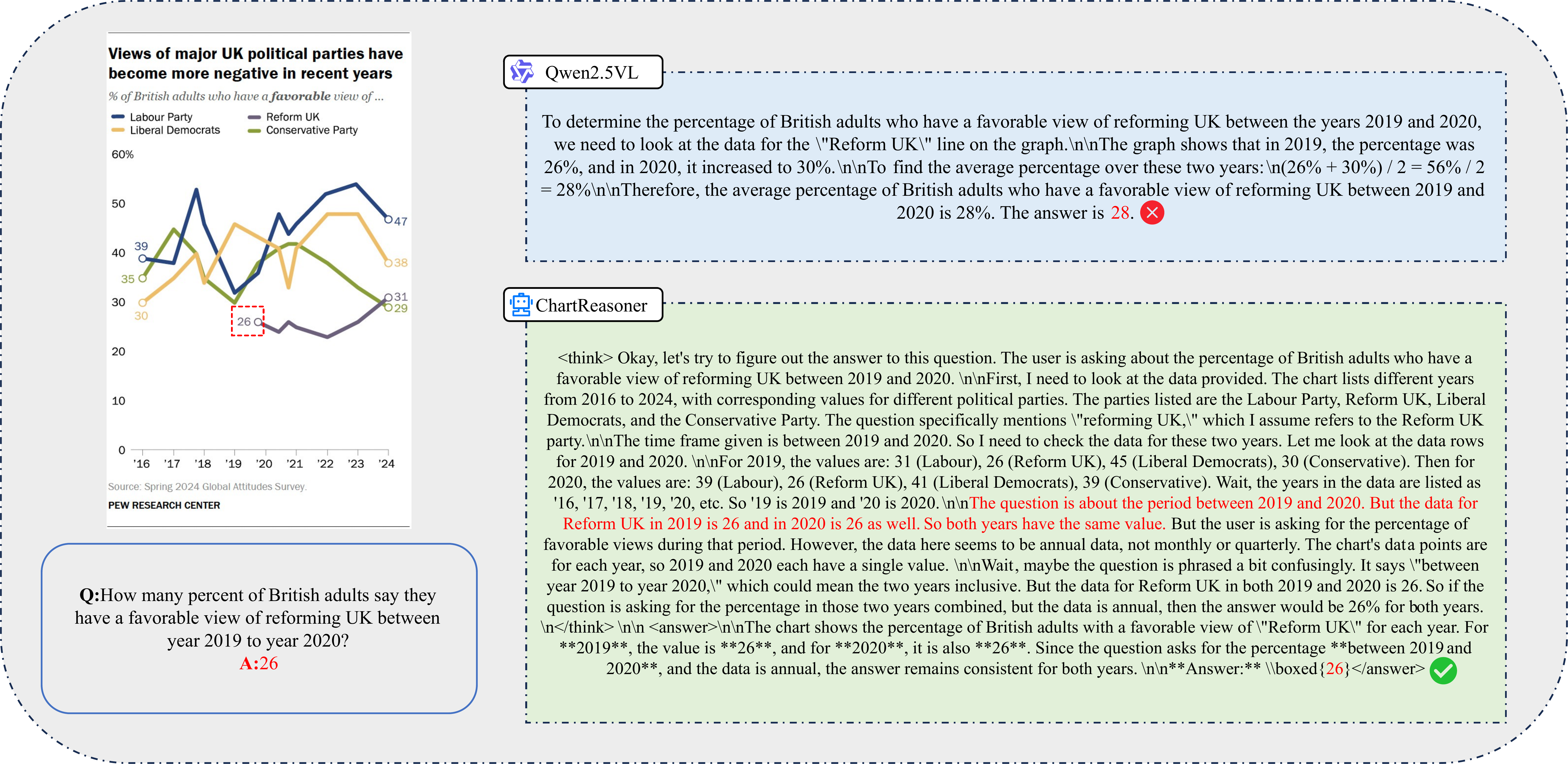}
    \caption{Comparison of Model Responses in ChartQA on a Line Chart.}
    \vspace{-3mm}
    \label{fig:Example13}
\end{figure*}

\begin{figure*}[t]
    \centering
    \includegraphics[width=\linewidth]{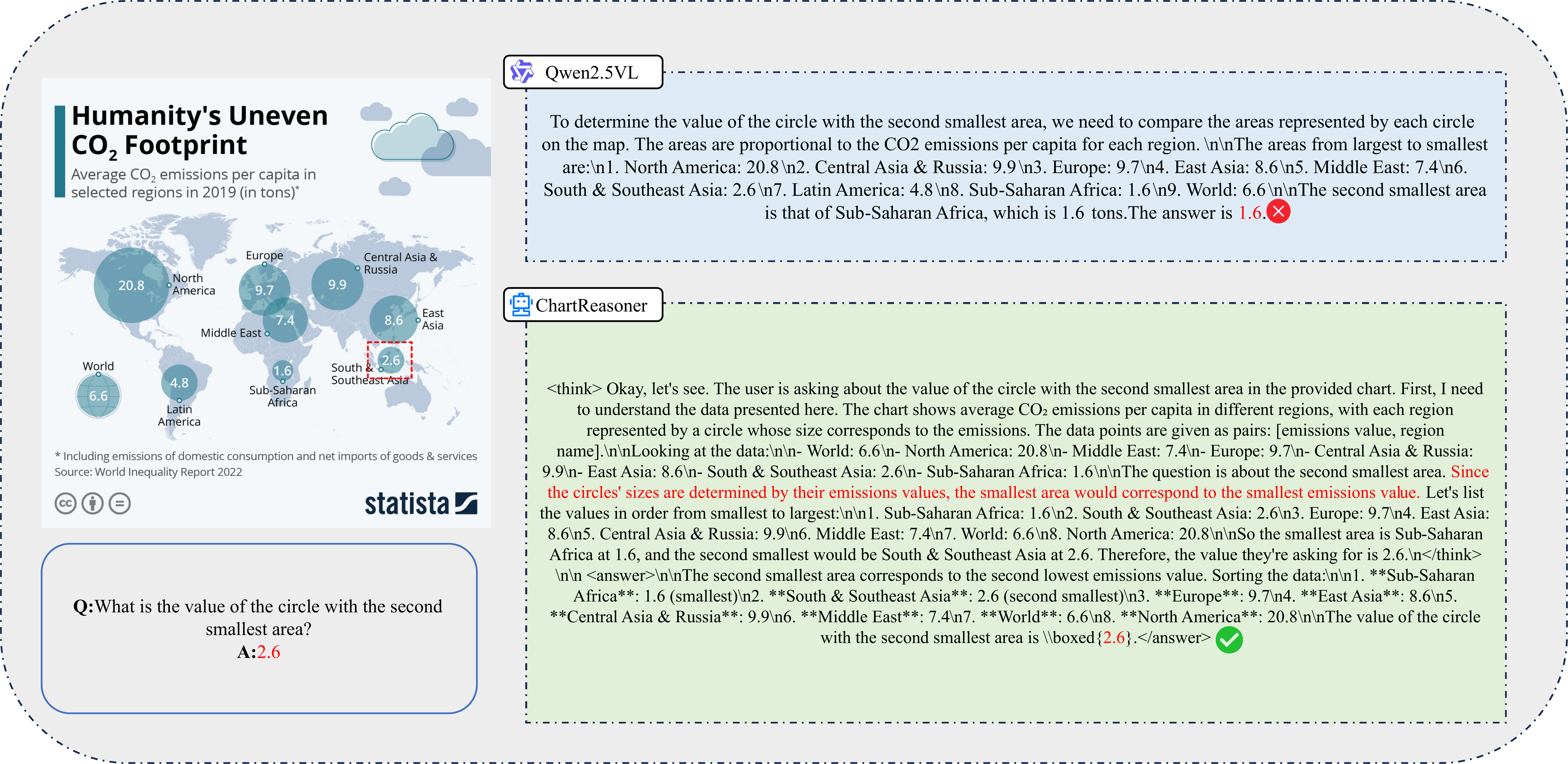}
    \caption{Comparison of Model Responses in ChartQA on a Scatter Chart.}
    \label{fig:Example4}
\end{figure*}

To further illustrate the performance improvements brought by our model in chart-based multimodal reasoning, we conduct a qualitative analysis using representative examples. These cases help demonstrate how enhanced reasoning capabilities can effectively assist visual understanding, especially when direct visual recognition is ambiguous or when the question requires complex logical interpretation. As illustrated in \autoref{fig:Example11}--\autoref{fig:Example4}, these examples further demonstrate the effectiveness of our method.

\paragraph{Visual-Aided Reasoning.}
One core strength of our ChartReasoner lies in its ability to perform visual reasoning that supplements and corrects potentially uncertain visual recognition. As shown in \autoref{fig:Example11}, the example question is: "What is the label of the highest bar of February?" This task requires the model to first locate February on the x-axis and then identify the label corresponding to its highest bar—thus constituting a visual reasoning problem.

While baseline models such as Qwen2.5VL fail to correctly locate "February" and incorrectly identify "Sales" as the highest category, ChartReasoner demonstrates a more accurate analysis by first reasoning through the axis structure: "The x-axis data is [January, February, March, ..., December], so February is the second month." This allows it to correctly localize the February column and extract the corresponding bar label, thereby arriving at the correct answer.

This example highlights that reasoning capabilities can effectively compensate for limitations in visual recognition, particularly when axis elements or data labels are densely packed, occluded, or ambiguously rendered.

\paragraph{Complex Semantic Reasoning.}
In addition to visual grounding, ChartReasoner also excels in handling complex semantic questions that require precise logical understanding. As shown in \autoref{fig:Example2}, the example question is: "How many percent of U.S. coffee drinkers drink less than 2 cups of coffee at home on a weekday?" The key to this question lies in correctly interpreting the condition "less than 2 cups." However, Qwen2.5VL incorrectly includes the "2 cups" category in its calculation, leading to a wrong answer. In contrast, ChartReasoner demonstrates its advanced reasoning by recognizing the logical boundary of the query and explicitly excluding the 2-cup group from its aggregation, yielding the correct answer. This indicates that reasoning ability is critical for precise comprehension of quantitative and conditional logic, which is often required in real-world ChartQA scenarios.

\section{Prompt Design for Visual Evaluation with GPT-4V}
\label{appendix:gpt_pro}
To comprehensively assess the visual quality of generated charts, we adopt a structured prompt-based evaluation approach using GPT-4V. The prompt instructs the model to compare a generated chart with its corresponding ground-truth version and assign a similarity score ranging from 1 to 10. The scoring is based on four key criteria: Colors (accuracy of color schemes), Axes \& Scale (consistency of axis ranges and units), Data Points Position (placement and alignment of bars, lines, or markers), and Overall Layout (correctness of titles, labels, legends, etc.).

\begin{figure*}[t]
    \centering
    \includegraphics[width=\linewidth]{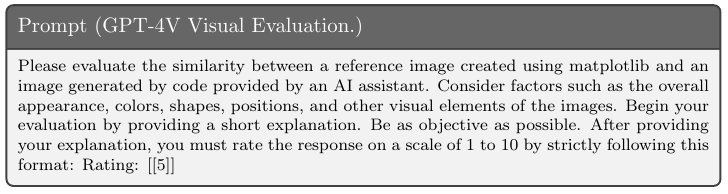}
    \caption{GPT-4V Visual Evaluation Prompt.}
    \label{fig:gpt4v-prompt}
\end{figure*}

This prompt enables GPT-4V to produce fine-grained visual judgments that go beyond traditional execution-based metrics (e.g., code correctness), capturing layout-level discrepancies that impact real-world interpretability. An example of such a prompt is illustrated in \autoref{fig:gpt4v-prompt}. This evaluation method bridges the gap between syntactic correctness and perceptual fidelity in chart generation tasks.




\section{Prompt Engineering for ECharts Code Generation}
\label{appendix:html_pro}
To enable effective chart generation, we employ domain-specific prompt engineering tailored to the ECharts visualization framework. The prompts are constructed to cover 18 thematic domains and 111 subtopics, spanning social, economic, technological, and environmental dimensions. This ensures diverse coverage of chart types and semantic contexts.

\begin{figure*}[t]
    \centering
    \includegraphics[width=\linewidth]{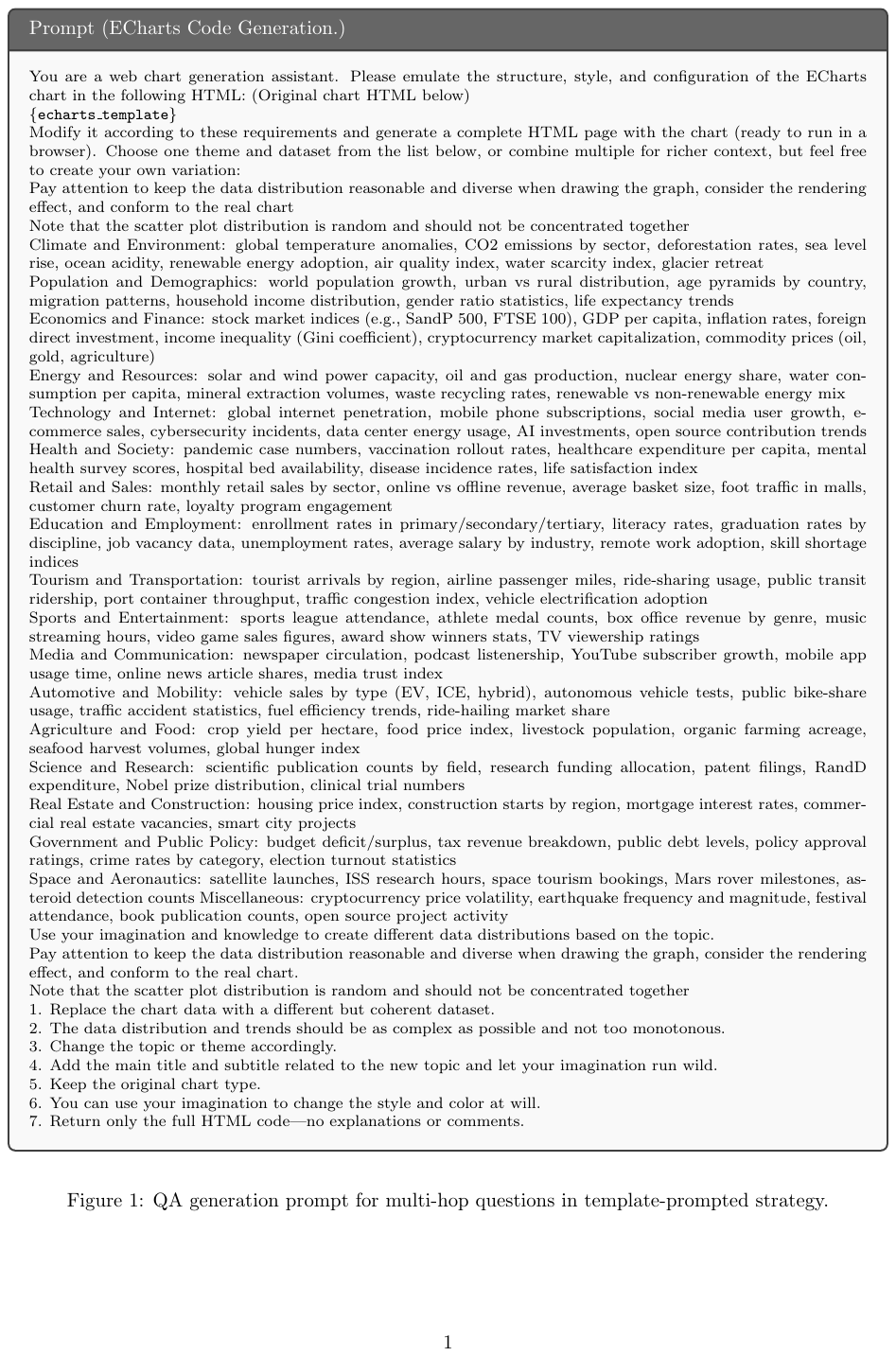}
    \caption{ECharts Code Generation Prompt.}
    \label{fig:gen_prompt}
\end{figure*}

Each prompt clearly specifies the chart topic, the intended visual form (e.g., bar chart, line chart, scatter plot), and any constraints on the layout or data encoding. As demonstrated in \autoref{fig:gen_prompt}, this guided prompting allows models like DeepSeek R1 to leverage their strong reasoning abilities to produce structurally varied and semantically rich visualizations. These prompts are essential to ensure that the generated charts are not only syntactically valid but also meaningful and domain-relevant.

\section{Chart-to-Code Dataset Detailed Case}
\label{appendix:c2c_case}
To further illustrate the design of our Chart-to-Code Dataset, we present selected examples that directly show the generated ECharts HTML code alongside the corresponding rendered chart. These examples also highlight the flexibility of the chart template system and the reasoning capability of the DeepSeek R1 model in generating structurally complex and thematically rich charts. By showcasing a range of chart types—including bar, line, and pie charts—these cases reflect the robustness of our prompt engineering approach and the effectiveness of the multi-stage quality filtering pipeline described in the methodology. \autoref{fig:html_Example1}--\autoref{fig:html_Example4} present more detailed examples from the Chart2Code dataset.

\begin{figure*}[h]
    \centering
    \includegraphics[width=0.8\linewidth]{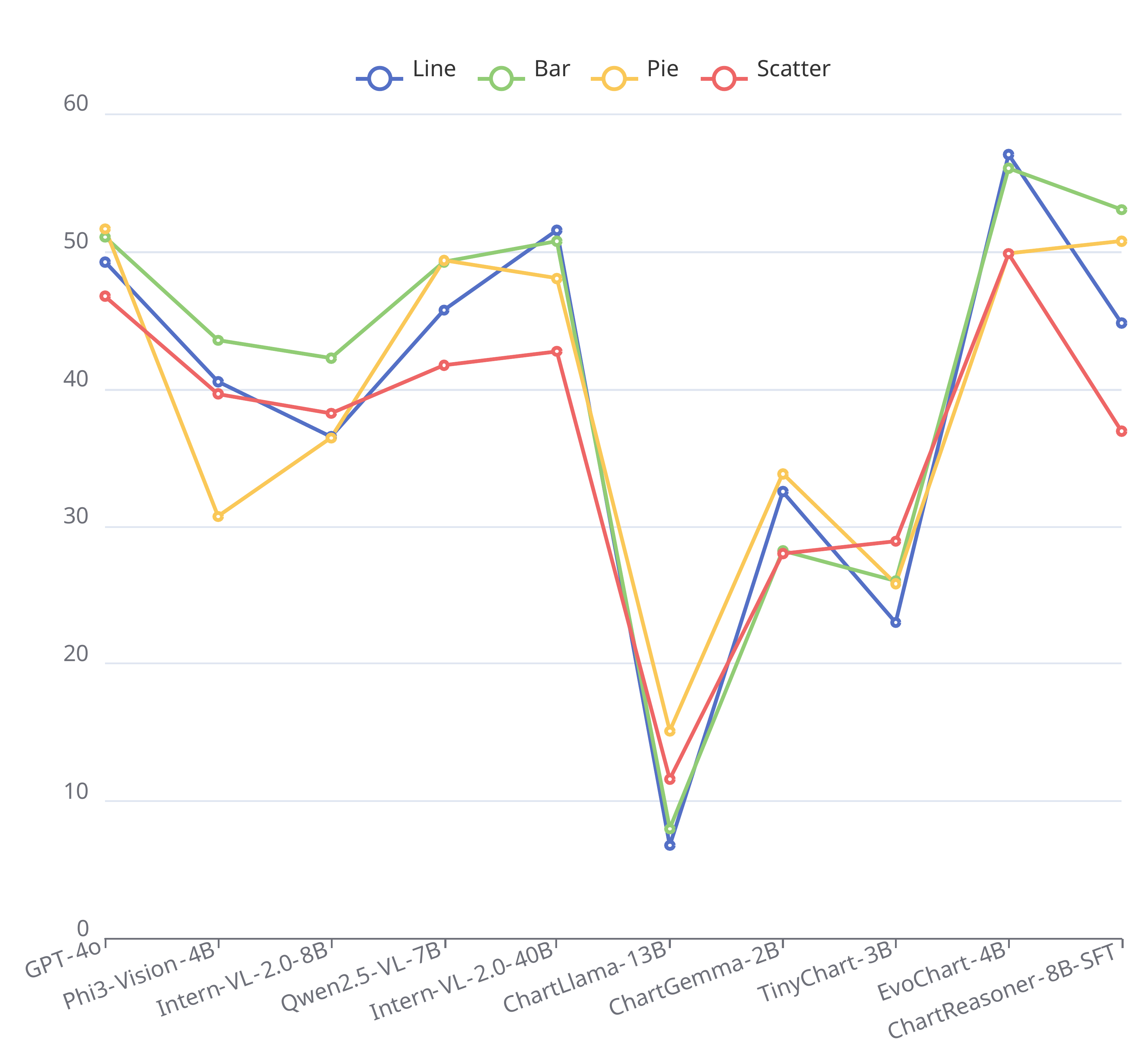}
    \caption{EvoChart Performance Across Chart Types.}
    \vspace{-3mm}
    \label{fig:EvoChart-charttype}
\end{figure*}

\begin{figure*}[t]
    \centering
    \includegraphics[width=\linewidth]{image/html_case1_new.pdf}
    \caption{Example 1 from the Chart-to-Code Dataset.}
    \label{fig:html_Example1}
\end{figure*}

\begin{figure*}[t]
    \centering
    \includegraphics[width=\linewidth]{image/html_case2_new.pdf}
    \caption{Example 2 from the Chart-to-Code Dataset.}
    \label{fig:html_Example2}
\end{figure*}

\begin{figure*}[t]
    \centering
    \includegraphics[width=\linewidth]{image/html_case3_new.pdf}
    \caption{Example 3 from the Chart-to-Code Dataset.}
    \label{fig:html_Example3}
\end{figure*}

\begin{figure*}[t]
    \centering
    \includegraphics[width=\linewidth]{image/html_case4_new.pdf}
    \caption{Example 4 from the Chart-to-Code Dataset.}
    \label{fig:html_Example4}
\end{figure*}

\end{document}